\documentclass[conference]{IEEEtran}
\IEEEoverridecommandlockouts
\usepackage{cite}
\usepackage{amsmath,amssymb,amsfonts}

\usepackage[titlenumbered,ruled]{algorithm2e}

\usepackage{comment}
\usepackage{algorithmic}
\usepackage{graphicx}
\usepackage{textcomp}
\usepackage{xcolor}
\usepackage{caption}
\usepackage{subcaption}
\usepackage{tabularx} 
\def\BibTeX{{\rm B\kern-.05em{\sc i\kern-.025em b}\kern-.08em
    T\kern-.1667em\lower.7ex\hbox{E}\kern-.125emX}}
\begin{document}

%\title{Context Normalization: A Local Normalization Technique for Image Pre-processing}

\title{Context Normalization Layer with Applications}

\author{\IEEEauthorblockN{1\textsuperscript{st} Bilal FAYE}
\IEEEauthorblockA{\textit{LIPN, UMR CNRS 7030}\\ 
\textit{Sorbonne Paris Nord University}\\
Villetaneuse, France \\
faye@lipn.univ-paris13.fr}
\and
\IEEEauthorblockN{2\textsuperscript{rd} Hanane AZZAG}
\IEEEauthorblockA{\textit{LIPN, UMR CNRS 7030}\\ 
\textit{Sorbonne Paris Nord University}\\
Villetaneuse, France \\
azzag@univ-paris13.fr}
\and
\IEEEauthorblockN{3\textsuperscript{th} Mustapha Lebbah}
\IEEEauthorblockA{\textit{DAVID Lab, University of Versailles,} \\
\textit{Universit\'e Paris-Saclay,}\\
Versailles, France \\
mustapha.lebbah@uvsq.fr}
\and
\IEEEauthorblockN{4\textsuperscript{nd} Mohamed-Djallel DILMI}
\IEEEauthorblockA{\textit{LIPN, UMR CNRS 7030}\\ 
\textit{Sorbonne Paris Nord University}\\
Villetaneuse, France \\
dilmi@lipn.univ-paris13.fr}
\and
\IEEEauthorblockN{5\textsuperscript{th} Djamel BOUCHAFFRA}
\IEEEauthorblockA{\textit{Center for Development of Advanced Technologies} \\
\textit{Algiers, Algeria}\\
djamel.bouchaffra@gmail.com}
}

\maketitle

\begin{abstract}
Deep neural networks (DNNs) have gained prominence in many areas such as computer vision (CV), natural language processing (NLP), robotics, and bioinformatics. While their deep and complex structure enables powerful representation and hierarchical learning, it poses serious challenges  (e.g., internal covariate shift, vanishing/exploding gradients, overfitting, and computational complexity), during their training phase. Neuron activity normalization is an effective strategy that lives up to these challenges. This procedure consists in promoting stability, creating a balanced learning, improving performance generalization and gradient flow efficiency. Traditional normalization methods often overlook inherent dataset relationships. For example, batch normalization (BN) estimates mean and standard deviation from randomly constructed mini-batches (composed of unrelated samples), leading to performance dependence solely on the size of mini-batches, without accounting for data correlation within these batches. Conventional techniques such as Layer Normalization, Instance Normalization, and Group Normalization estimate normalization parameters per instance, addressing mini-batch size issues. Mixture Normalization (MN) utilizes a two-step process: (i) training a Gaussian mixture model (GMM) to determine components parameters, and (ii) normalizing activations accordingly. MN outperforms BN but incurs computational overhead due to GMM usage.
To overcome these limitations, we propose a novel methodology that we named "Context Normalization" (CN). Our approach assumes that the data distribution can be represented as a mixture of Gaussian components. However, unlike MN that assumes a-priori that data are partitioned with respect to a set of Gaussian distributions, CN introduces the notion of concept that accounts for data relationship via a neural network classification scheme. Samples that are gathered within a cluster define a context. The estimation of the Gaussian components parameters is conducted through a supervised neural network-based concept classification. CN is more precise when clusters are thick and not sparse. Extensive comparative experiments conducted on various datasets demonstrates the superiority of CN over BN and MN in terms of convergence speed and performance generalization. In fact, CN outperforms BN and MN with a convergence speed margin of 5\% and a performance margin of 10\%. These results reveal the importance and the need of capturing inherent data context to learn the Gaussian component parameters. Our proposed approach harnesses data relationships, and therefore enhances deep learning models in various applications.
\end{abstract}

\section{Introduction}
Normalization is a general process that transforms data to possess specific statistical properties. Various methods exist for data normalization, with some of the most common techniques including centering, scaling, standardizing, decorrelating, and whitening~\cite{kessy2018optimal}. Input normalization can be used in deep neural networks (DNNs) training to remove the magnitude difference between features, speeding convergence during linear model optimization~\cite{lecun2002efficient}. In layered neural networks, as the input is only directly connected to the first weight matrix, it is not clear how the input impacts the optimization landscape with respect to other weight matrices. The initial weights are typically not normalized, and this can indeed have a significant impact on the gradient optimization procedure. To overcome this, some methods employ weight-initializing techniques to obtain equal variances for layer input/output gradients across different layers~\cite{he2015delving}. However, due to the updating of the weight matrices during training, the equal variance property across layers may be broken. From this perspective, it is important to normalize activations during training, to obtain similar benefits of normalizing inputs.\newline
Different normalization techniques, including activation normalization, weight normalization, and gradient normalization, are employed to enhance the training performance of DNNs.
To normalize activations, the most common technique is Batch Normalization (BN)~\cite{ioffe2015batch}. BN has been proposed to solve the problem caused by the changing distribution of the inputs of each layer during training, called internal covariate shift. In fact, the distribution change problem often requires lower learning rates and careful parameter initialization, which slows down the training process and makes it difficult to train models with saturating nonlinearity. As a layer of the neural network, BN methods perform the normalization on each training mini-batch with its respective mean and variance. This mini-batch-wise approach allows for a more appropriate representation of the data and, therefore, faster processing, thus increasing the performance of the neural network in terms of convergence. Despite the good performance, the effect of BN depends on the mini-batch size, and it is not obvious how to apply it to some DNNs architectures. To address this, several variants and alternative methods have been proposed~\cite{huang2020normalization}.\newline
BN and its few extensions can be studied from the viewpoint of Fisher kernels that arise from generative probability models. Kalayeh and al.~\cite{kalayeh2019training} show that assuming samples within a mini-batch are from the same probability density function, then BN is identical to the Fisher vector of a Gaussian distribution. More specifically, each instance in the mini-batch is assigned to a component of the Gaussian Mixture Model (GMM), where the GMM approximates the probability density function of the input activations. Instead of computing one set of statistical measures from the entire population (of instances in the mini-batch) as BN does, Mixture Normalization (MN)~\cite{kalayeh2019training} proposes a normalization on sub-populations which can be identified by disentangling modes of the distribution, estimated via GMM. In addition to speeding up training, MN allows better accuracy results to be achieved.\newline
\indent In this perspective, we propose a novel normalization technique called context normalization (CN). In fact, assuming that the data are well modeled by a mixture of several components, each sample in the mini-batch is normalized using the mean and variance of the associated component. Indeed, the capability of GMM to approximate any continuous distribution with arbitrary precision has been demonstrated by~\cite{titteringtonf}. Building upon this foundation, our paper follows a similar track but introduces a novel method. In particular, we define a context that can come from various sources that describe the structure of the dataset. A context can be conceptualized as a coherent cluster of samples that share common characteristics and can be effectively grouped together. Each context can be viewed as a component of the Gaussian mixture with its own probability density function. By normalizing samples from the same context with the parameters learned during backpropagation, CN allows an estimation of the mean and variance of each mixture component thus improving the discrimination power of the data representation according to the target task.\newline
\indent In summary, the main contributions of this work are as follows:
\begin{itemize}
\item We propose Context Normalization (CN), a novel approach that utilizes defined contexts to capture underlying distribution variations. In CN, each sample in a mini-batch is normalized using the mean and standard deviation specific to its context. By treating contexts as components of a Gaussian mixture, we learn their parameters during model training, eliminating the need for the EM algorithm. This leads to improved efficiency and simplified implementation of CN.

    \item Through a comprehensive set of experiments, we demonstrate that CN not only accelerates model convergence, but also achieves superior final test accuracy. These results highlight the effectiveness of our proposed method in improving the overall performance of models.
\end{itemize}
For consistency, we use the variable notation proposed in Table~\ref{table:notations} for all sections.
\begin{table}[!ht]
    \centering
    \begin{tabular}{|l|c|}
        \hline
        \textbf{Variable}  & \textbf{Definition} \\
        \hline
        $T$ & number of context \\
        \hline
        $K$ & number of components \\
        \hline
        $N$ & mini-batch size\\
        \hline
        $C$ & channels\\
        \hline
         $H$ & height \\
        \hline
        $W$ & width \\
        \hline
       
        $Net$ & neural network \\
        \hline
        $\Theta$ &  trainable parameters of neural network \\
        \hline
        $x \in \mathbb R^{N\times C\times H\times W}$ & 4-D activation tensor  \\
        \hline
       % $B = \{x_{1...m}\}$ & set of elements (mini-batch) \\
       $B = \{x_{1:m}\}$ & flattened $x$ across axis $N$, $H$ and $W$ \\
        \hline
        $B_i$ & flattened $x$ across axis $H$ and $W$ \\
        \hline
       $\mu_r  $ & the mean on the context $r$
        \\
        \hline
        $\sigma_r   $ &standard deviation on the context $r$
        \\
        \hline
         $onehot(.)  $ & One-Hot Encoding function
        \\
        \hline
    \end{tabular}
    \caption{Table of notations}
    \label{table:notations}
\end{table}

\section{RELATED WORK}
\subsection{Batch Normalization}\label{section:batch_normalization}
Let $x \in \mathbb R^{N\times C\times H\times W}$ denote the activation for a given neuron in a layer of a convolutional neural network where $N$, $C$, $H$ and $W$ are respectively the batch, channel, height, and width axes. Batch normalization (BN)~\cite{ioffe2015batch} standardizes with $m$ samples mini-batch 
$B = \{x_{1:m}: m \in [1, N]\times [1, H] \times [1, W]\}$
with flattened $x$ across all but the channel axis by:
\begin{equation}
\label{bn_equation}
    \hat{x}_{i} = \frac{x_{i}-\mu_B}{\sqrt{\sigma^2_B+\epsilon}},
\end{equation}
where $\mu_B = \frac{1}{m} \sum_i^m x_{i}$, $\sigma^2_B = \frac{1}{m}\sum_i^m (x_{i}-\mu_B)^2$ are the mean and variance respectively, and $\epsilon > 0$ a small number to prevent numerical instability.\newline
Due to the constraints introduced by standardization, additional learnable parameters $\gamma$ and $\beta$ are introduced to eliminate the linear regime of nonlinearity of some activations:
\begin{equation}
    \Tilde{x}_{i} = \gamma\hat{x}_{i} + \beta
\end{equation}
During inference, population statistics are needed for deterministic inference. They are usually computed by running the average over the training iterations, as follows:
\begin{equation}
    \begin{cases}
        \hat{\mu} = (1-\lambda)\hat{\mu} + \lambda \mu_B \\
        \hat{\sigma}^2 = (1-\lambda)\hat{\sigma}^2 + \lambda \sigma^2_B
    \end{cases}
\end{equation}
If the samples within the mini-batch are from the same distribution, the transformation in Equation~\eqref{bn_equation} generates a zero mean and unit variance distribution. This zero-mean and unit-variance constraint allows stabilizing the distribution of the activations and thus benefits training.\newline
\indent This mini-batch-wise approach makes it possible to have a more suitable representation of the data and, therefore, faster processing, thus increasing the performance of the neural network in terms of convergence. Despite the good performance, the effect of BN is dependent on the mini-batch size, and
the discrepancy between training and inference limits its usage in complex networks (e.g. recurrent neural networks). To tackle these challenges and address parameter estimation on unrelated samples, numerous variants, and alternative methods have been proposed.
\subsection{Variants of Batch Normalization}\label{section:general}
Some extensions to batch normalization (BN) have been proposed, in particular, Layer Normalization (LN), Instance Normalization (IN), Group Normalization (GN) and Mixture Normalization (MN)~\cite{huang2020normalization,kalayeh2019training}.\newline
The general transformation $x \rightarrow \hat{x}$ according to a mini-batch, on the flattened spatial domain $(L=H\times W)$ , can be written as follows
\begin{equation}
\label{general_transform}
    v_{i} = x_{i} - \mathbb{E}_{B_i}(x), \ 
    \hat{x}_{i} = \frac{v_{i}}{\sqrt{\mathbb{E}_{B_i}(v^2)+\epsilon}},
\end{equation}
given $B_i = \{j: j_N \in [1, N], j_C \in [i_C], j_L \in [1, L]\}$,
where $i = (i_N, i_C, i_L)$ a vector indexing the activations $x \in \mathbb R^{N\times C \times L}$.\newline
\textbf{Layer Normalization (LN)} eliminates inter-dependency of batch-normalized activations by calculating mean and variance based on specific layer neuron inputs. LN is effective for recurrent networks but may face challenges with convolutional layers due to variations in visual information across the spatial domain.\newline
\textbf{Instance Normalization (IN)} is a normalization technique that normalizes each sample individually, focusing on removing style information, especially in images. By computing mean values and standard deviations in the spatial domain, IN improves the performance of specific deep neural networks (DNNs) and finds widespread application in tasks like image style transfer~\cite{dumoulin2016learned}~\cite{huang2018multimodal}.\newline
\textbf{Group Normalization (GN)} is a normalization technique that divides neurons into groups and independently standardizes layer inputs for each sample within the groups. It excels in visual tasks with limited batch sizes, like object detection and segmentation.\newline
\textbf{Mixture Normalization}
In the context of deep neural networks (DNNs), the distribution of activations is almost certain to have multiple modes of variation due to the non-linearities. The batch normalization (BN)~\cite{ioffe2015batch} hypothesis that a Gaussian distribution can model the generative process of mini-batch samples is less valid. To address this, Mixture Normalization (MN)~\cite{kalayeh2019training} investigates BN from the viewpoint of Fisher kernels, which arise from generative probability models. Rather than using a mean and standard deviation calculated over entire instances within a mini-batch, MN uses a Gaussian Mixture Model (GMM) to affect each instance in the mini-batch to a component and then normalizes with respect to multiple means and standard deviations associated with different modes of variation in the underlying data distribution.\newline
\indent MN normalizes each sample in the mini-batch using the mean and standard deviation of the mixture component to which the sample belongs to. The probability density function $p_\theta$ that describes the data can be parameterized as a Gaussian Mixture Model (GMM). Let $x$ in $\mathbb{R}^D$, if $\theta = \{\lambda_k, \mu_k, \Sigma_k: k = 1, ..., K\}$,
\begin{equation}
\label{gmm}
    p(x) = \sum_{k=1}^K \lambda_kp(x|k),\  s.t.\ \forall_k\ :\ \lambda_k\ \ge 0,\   \sum_{k=1}^{K}\lambda_k=1,
\end{equation}
where
\begin{equation*}
    p(x|k)  = \frac{1}{(2\pi)^{D/2}\lvert\Sigma_k\rvert^{1/2}}\exp\left(-\frac{( x-\mu_k)^T\Sigma_k^{-1}( x-\mu_k)}{2}\right),
\end{equation*}
is the $k^{th}$ Gaussian in the mixture model $p(x)$, $\mu_k$ the mean vector and $\Sigma_k$ is the covariance matrix.\newline
The probability that $x$ has been generated  by the $k^{th}$ Gaussian component in the mixture model can be defined as:
\begin{equation*}
    %\label{mn_aggragation}
    \tau_k(x) = p(k|x) = \frac{\lambda_k p(x|k)}{\sum_{j=1}^K\lambda_j p(x|j)},
\end{equation*}
Based on these assumptions and the general transform in Equation~\eqref{general_transform}, the Mixture Normalizing Transform for a given $x_i$ is defined as
\begin{equation}
    \label{mn_aggregation}
    \hat{x}_{i} = \sum_{k=1}^K \frac{\tau_k(x_{i})}{\sqrt{\lambda_k}}\hat{x}_{i}^k,
\end{equation}
given
\begin{equation}
    \label{mn_norm}
    v_{i}^k=x_{i} - \mathbb E_{B_i}[\hat{\tau}_k(x).x], \  
    \hat{x}_{i}^k = \frac{v_{i}^k}{\sqrt{\mathbb E_{B_i}[\hat{\tau}_k(x).(v^k)^2]+\epsilon}},
\end{equation}
where
\begin{equation*}
    \hat{\tau}_k(x_{i}) = \frac{\tau_k(x_{i})}{\sum_{j \in B_i} \tau_k(x_{j})},
\end{equation*}
is the normalized contribution of $x_i$ with respect to the mini-batch $B_i$ in the estimation of the statistical measures of the $k^{th}$ Gaussian component.\newline
With this approach, Mixture Normalization can be done in two stages:
\begin{itemize}
    \item estimation of mixture model's parameters $\theta = \{\lambda_k, \mu_k, \Sigma_k: k = 1, ..., K\}$ by Expectation-Maximization (EM)~\cite{Dempster77maximumlikelihood} algorithm.
    \item normalization of each sample with respect to the estimated parameters (Equation~\eqref{mn_norm}) and aggregation using posterior probabilities (Equation~\eqref{mn_aggregation}).
\end{itemize}
On convolutional neural networks, this method allows Mixture Normalization to achieve better results than batch normalization in terms of convergence and accuracy in supervised learning tasks.
\section{Proposed Method: Context Normalization}
%In this section, based on the MN hypothesis, we will show how the use of GMM for sub-population identification and normalization with respect to not one but several components can improve BN and how learning the parameters of each component during training can give better results.\newline
%\indent Based on Mixture Normalization (MN) hypothesis ~\cite{kalayeh2019training} (ref. Figure~\ref{fig:all}), our Context Normalization (CN) method makes the same assumption that the data can be well represented by a mixture of several components instead of one (as in batch normalization (BN)~\cite{ioffe2015batch}). 
%
\subsection{Method description}
\begin{figure}[htbp]
\centering
%[width=.5\linewidth, height=.5\linewidth]
\includegraphics[ height=.5\linewidth]{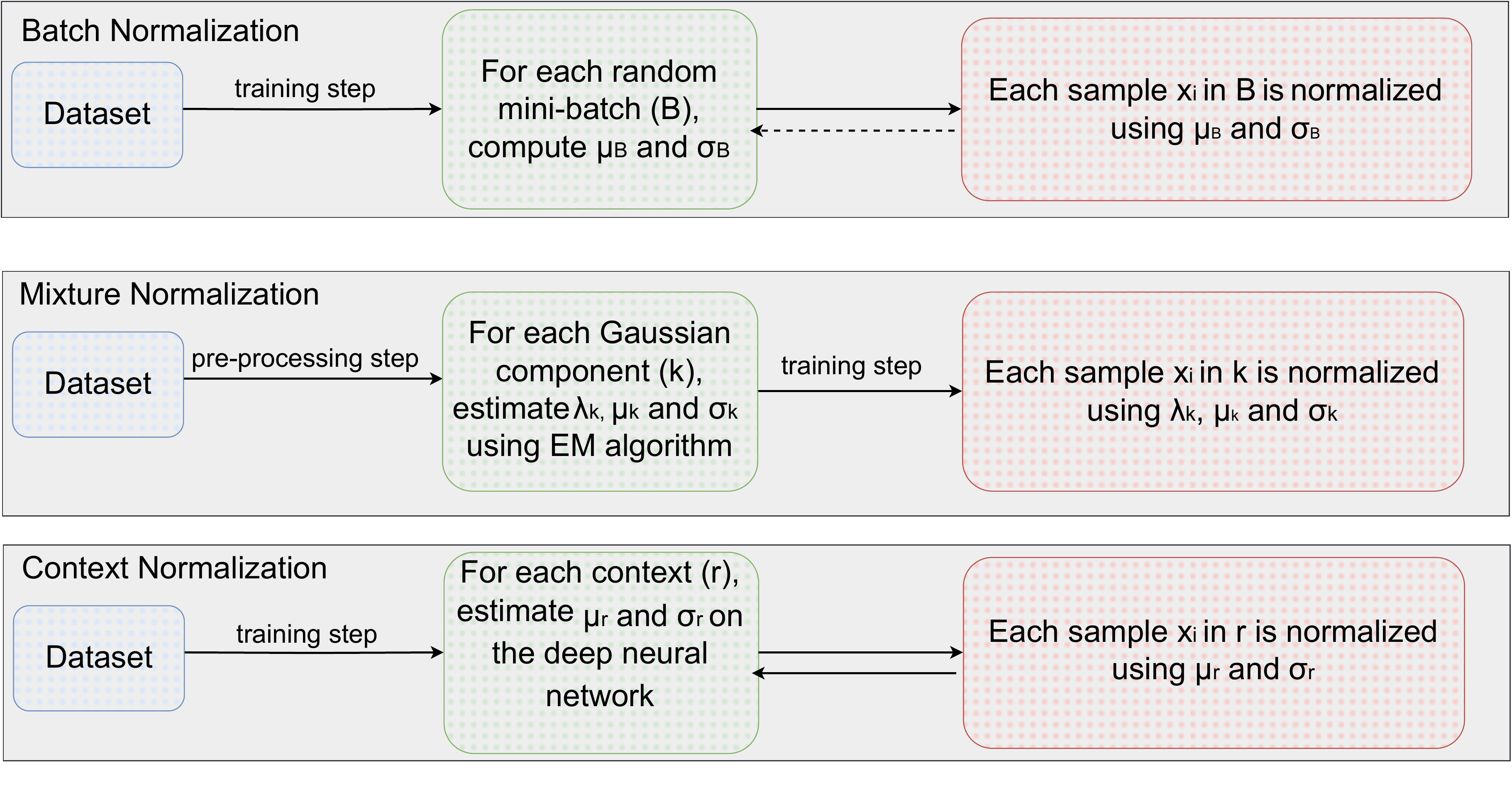}
\caption{
A concise overview of the processing steps involved in Batch Normalization (BN), Mixture Normalization (MN), and Context Normalization (CN). The dashed line in the Batch Normalization diagram indicates a mini-batch parameter update, highlighting a key step in the process.
}
\label{fig:all}
\end{figure}

Based on the Mixture Normalization (MN) hypothesis proposed by~\cite{kalayeh2019training} (ref. to Figure~\ref{fig:all}), our Context Normalization (CN) approach operates under a similar assumption that data can be effectively represented by a mixture of multiple components, as opposed to batch normalization (BN) ~\cite{ioffe2015batch}).  In the Context Normalization (CN) approach, a fundamental concept is introduced, namely, the notion of context, which represents a cluster of samples sharing common characteristics that can be efficiently grouped together. Unlike the Expectation-Maximization (EM) algorithm~\cite{Dempster77maximumlikelihood} typically employed for parameters estimation in each component, CN utilizes a deep neural network to learn these parameters through context-based normalization.
In our approach, we assign a unique identifier to each context and utilize it for normalization during training. Samples within the same context share the same identifier, allowing for alignment in a shared space that aligns with the target task. This approach not only facilitates the normalization of samples within the same context but also enables the estimation of optimal parameters for all contexts, promoting the convergence of the model. By leveraging these context identifiers, our approach enhances the alignment and adaptability of the model to different contexts, leading to improved performance.\newline
\subsection{Parameters learning}
\begin{figure}[!h]
\centering
%[width=.5\linewidth, height=.5\linewidth]
\includegraphics[ height=.6\linewidth]{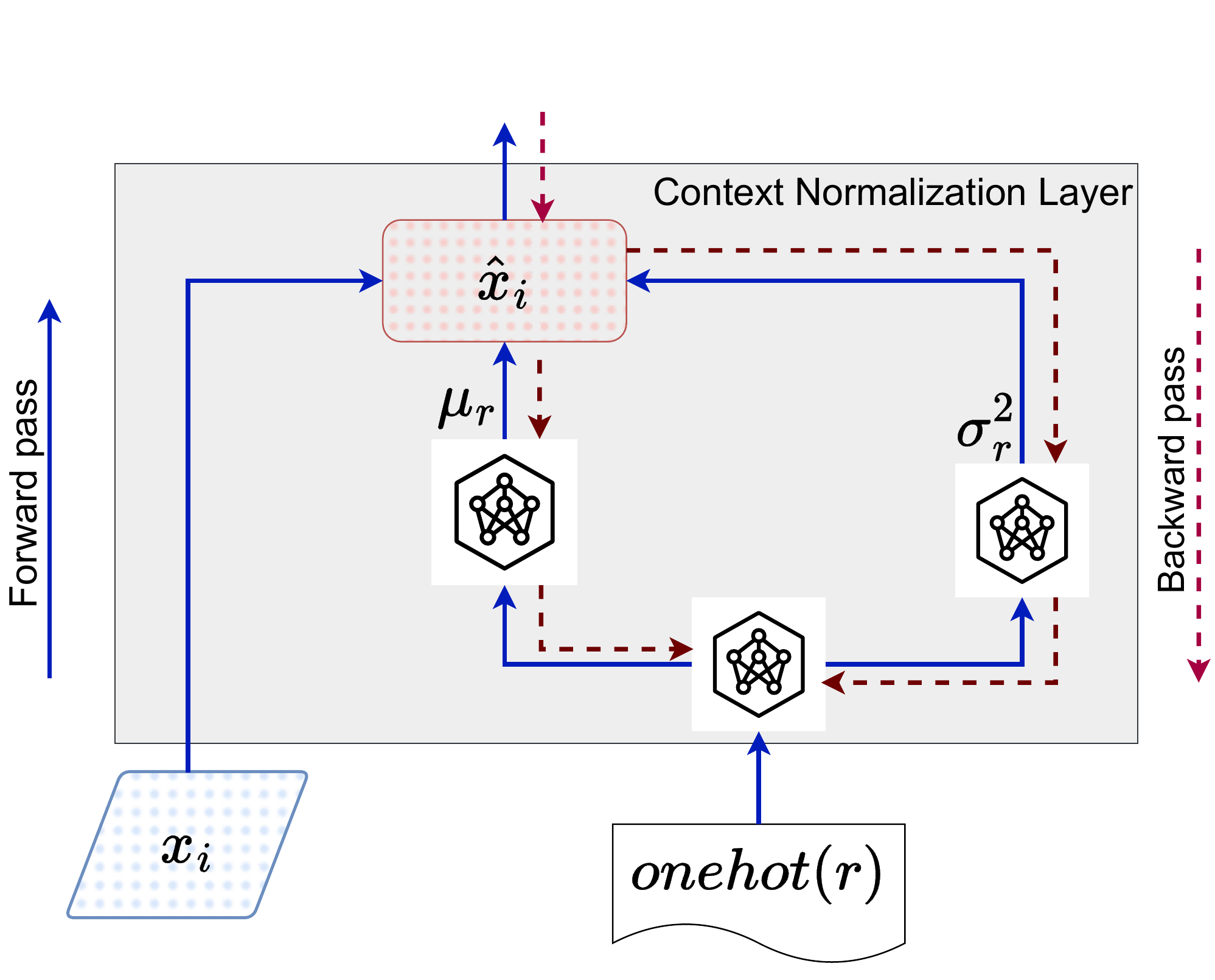}
\caption{Context Normalization Layer applied to a given activation $x_i$. The context identifier ($r$) is encoded by a neural network, the output of which is then used as input to two different neural networks to generate a mean ($\mu_r$) and a standard deviation ($\sigma_r$), respectively, for normalizing $x_i$.}
\label{fig:cn_layer}
\end{figure}

\indent For a given $x_i$ in $B_i$, the mean ($\mu_r$) and standard deviation ($\sigma_r$) are estimated based on the context $r$ associate to $x_i$. This estimation process is outlined in Algorithm~\ref{alg:one} and visually depicted in Figure~\ref{fig:cn_layer}. Subsequently, these estimated parameters are used to normalize $x_i$ according to Equation~\eqref{equation:cn_nom}. The CN transform can be added to a deep neural network to provide a better representation of the input data. In Algorithm~\ref{alg:one}, $\hat{x}_i = CN(x_i, r)$ indicates that the parameters $\mu_r$ and $\sigma_r$ are to be learned depending on each training example according to the context $r$. Algorithm~\ref{alg:two} summarizes the procedure for training context-normalized networks.\newline
\begin{equation}\label{equation:cn_nom}
    \hat{x}_i \gets \frac{x_i - \mu_r}{\sqrt{\sigma^2_r+\epsilon}}
\end{equation}
\begin{algorithm}
\caption{CN- Context Normalizing Transform}\label{alg:one}
\SetKwInOut{KwIn}{Input}
\SetKwInOut{KwOut}{Output}

\KwIn{Values of $x$ over the mini-batch $B_i=\{x_i\}_{i=1}^m$; context $r$ associated to $x_i$}
\KwOut{\{$\hat{x_i}$ = CN($x_i$, $r$)\}}
%\If{$x$ and $cv$ are not empty}{
 % $\hat{r} 
  $ \alpha_r \gets Embedder_{r}(onehot(r))$ //
 %$  \hat{c} \gets Embedder_{c}(c)$ // 
 {\small \it identifier embedding} \newline
 
  %  $\mu_r \gets Embedder_\mu(\hat{r})$ // {\small \it mean estimation }\newline
   
   $\mu_r \gets Embedder_\mu(\alpha_r )$ // {\small \it mean estimation }\newline
   
  %  $\sigma_r^2 \gets Embedder_\sigma(\hat{r})$ //{\small \it variance estimation} \newline
 $\sigma_r^2 \gets Embedder_\sigma(\alpha_r )$ // {\small \it variance estimation} \newline
 
$\hat{x}_i$ = CN$(x_i, r)$: Normalize $x_i$ using Equation~\ref{equation:cn_nom}
%}
\end{algorithm}

\begin{algorithm}
\caption{CN-Training (Context-Normalized Deep Neural Network)}\label{alg:two}
\SetKwInOut{KwIn}{Input}
\SetKwInOut{KwOut}{Output}

\KwIn{Deep neural network $Net$ with trainable parameters $\Theta$; subset of activations $B_i=\{x_i\}_{i=1}^m$ 
}
\KwOut{Context-Normalized deep neural network for training, $Net^{tr}_{CN}$}
    $Net^{tr}_{CN} = Net$ // {\small \it Training CN deep neural network }\\ 
    \For{$i \gets 1$ to $m$}{
        \begin{itemize}
            \item Add Algorithm~\ref{alg:one} transformation: $\hat{x}_i = $ CN$(x_i, r)$ \\ // {\small \it normalize $x_i$ using associated context, $r$}\\ 
            \item Modify each layer in $Net^{tr}_{CN}$ with $x_i$ to take $\hat{x}_i$\\ instead 
        \end{itemize}
    }
    Train $Net^{tr}_{CN}$ to optimize the parameters: $\Theta = \Theta \cup \{\mu_r, \sigma_r\}_{r=1}^T$ \\ 
\end{algorithm}
In algorithm~\ref{alg:one}, we use 
an affine transformation followed by element-wise linearity:
%\begin{equation}\label{equation:update}
  %  \hat{c} = \textit{W}_c.c + b_c,\ 
  %  \mu_c = \textit{W}_{mu}.\hat{c} + b_{\mu}, \ 
  %  \sigma_c^2 = \textit{W}_{\sigma}.\hat{c} + b_{\sigma},
%\end{equation}
$$ \alpha_r  \gets Embedder_{r}(onehot(r)): \textit{W}_r  \mbox{ } onehot(r) + b_r,$$
$$\mu_r \gets Embedder_\mu(\alpha_r) :   \textit{W}_{\mu} \mbox{ } \alpha_r  + b_{\mu},$$
$$ \sigma_r^2 \gets Embedder_\sigma(\alpha_r): \textit{W}_{\sigma} \mbox{ } \alpha_r  + b_{\sigma} $$

where $\textit{W}$ and $b$  are learned parameters of the {Embedder} model and $r$ is the context associated to $x_i$. The $onehot(.)$ is the function that performs one-hot encoding on the categorical variable representing the context $r$, which is embedded in the continuous space  $\alpha_r$.
\newline
During the training process, it is necessary to backpropagate the gradient of the loss function, denoted as $\ell$, through the transformation. Additionally, the gradients with respect to the parameters of the CN transform need to be computed. This is achieved using the chain rule, as shown in the following expression (before simplification):\newline
$$ \frac{\partial \ell}{\partial \mu_r} = \frac{\partial \ell}{\partial \hat{x}_i}.\frac{\partial \hat{x}_i}{\partial \mu_r} = -\frac{\partial \ell}{\partial \hat{x}_i}.(\sigma_r^2+\epsilon)^{-1/2}$$
$$\frac{\partial \ell}{\partial \sigma_r^2} = \frac{\partial \ell}{\partial \hat{x}_i}.\frac{\partial \hat{x}_i}{\partial \sigma_r^2} = \frac{\mu_r + x_i}{2(\sigma_r^2 + \epsilon)^{3/2}}$$
CN transform is a differentiable operation in deep neural networks that normalizes input data. By applying CN, the model can continuously learn from input distributions and adapt its representations to the target task, leading to improved performance. This normalization helps mitigate the influence of variations in input distributions, allowing the model to focus on relevant patterns and features. The differentiability of CN enables efficient gradient flow during training, facilitating parameter updates and learning from the normalized data while preserving differentiation through the normalization process. Overall, CN plays a vital role in enhancing model performance by promoting effective learning and adaptability through data normalization. It demonstrates higher flexibility compared to MN due to its ability to establish consistent data grouping based on provided contexts, without the need for additional algorithms. This is advantageous over MN since the Expectation-Maximization (EM) algorithm employed in MN can exhibit slower convergence. In the specific case of classifying dog images, where data scarcity is a challenge, the method addresses this issue by partitioning the dog class into subclasses. This approach enables the acquisition of specific features applicable to all dogs, facilitating the normalization of images within the dog superclass and creating a more coherent and easily learnable feature space. Importantly, the context identifier used for learning the normalization parameters is unrelated to the images themselves. Instead, it can be viewed as noise, contributing to the regularization of the deep neural network during training, similar to techniques like dropout, thereby enhancing the generalization performance of the model~\cite{noh2017regularizing}.

%Once the model has converged, we are in the same situation as mixture normalizing, with the parameters of each context known and dependent on the target task.

\indent After the model has converged, we enter a scenario similar to mixture normalization, where the parameters of each context are known and dependent on the target task. During the inference step, we have the option to normalize the data directly using CN or consider all the contexts collectively. This enhanced approach is referred to as CN+ (ref. Algorithm~\ref{alg:three}). For a given $x_i$ in $B_i$ and using the Equation~\ref{mn_aggregation}, CN+ can be formulated as
\begin{equation}
    \label{cn_aggregation}
  %  \hat{x}_{i} = \sum_{t=1}^C \frac{\tau_c(x_{i})}{\sqrt{\frac{1}{C}}}\hat{x}_{i}^c = \sqrt{C}.\sum_{t=1}^C \tau_c(x_{i})\hat{x}_{i}^c ,
     \hat{x}_{i} = \sqrt{T}\sum_{r=1}^{T} \tau_r(x_{i})\hat{x}_{i}^r ,
\end{equation}
given
\begin{equation*}
    \label{cn_norm}
    v_{i}^r=x_{i} - \mathbb E_{B_i}[\hat{\tau}_r(x).x], \ 
    \hat{x}_{i}^r = \frac{v_{i}^r}{\sqrt{\mathbb E_{B_i}[\hat{\tau}_r(x).(v^r)^2]+\epsilon}},
\end{equation*}
where
\begin{equation*}
    \hat{\tau}_r(x_{i}) = \frac{\tau_r(x_{i})}{\sum_{j \in B_i} \tau_r(x_{j})},
\end{equation*}
under the assumption that prior probabilities ($\lambda_r = \frac{1}{T}, r=1, ..., T$) are constant;
\begin{algorithm}
\caption{CN-Inference (Context-Normalized Deep Neural Network)}\label{alg:three}
\SetKwInOut{KwIn}{Input}
\SetKwInOut{KwOut}{Output}
\KwIn{Deep Neural Network $Net^{tr}_{CN}$ with trainable parameters $\Theta \cup \{\mu_r, \sigma_r\}_{r=1}^T$ (ref. Algorithm~\ref{alg:two}); mini-batch of activation $\{x_i\}_{i=1}^m$; $choice \in$ \{CN, CN+\}}
\KwOut{Context-Normalized deep neural network for inference, $Net^{inf}_{CN}$}
    $Net^{inf}_{CN} \gets Net^{tr}_{CN}$ // {\small \it Inference CN deep neural network}\\ //{\small \it with frozen parameters}\\
    \If{$choice=$ CN}{
        \For{$i \gets 1$ to $m$}{
            Normalize using context($r$) parameters ($\mu_r$ and $\sigma_r$ learned during training) associated to $x_i$.\\
            $\hat{x}_i \gets $ CN$(x_i, r)$ 
        }
    }
    \If{$choice=$ CN+}{
        \For{$i \gets 1$ to $m$}{
            \begin{itemize}
                \item Normalize $x_i$ using Equation~\ref{cn_aggregation} // \small \it Normalization + aggregation
            \end{itemize}
        }
    }
\end{algorithm}
\section{Experiments}
In the proposed experiments, we will assess the performance of the context normalization method and compare it to BN and MN. Detailed descriptions of the methods under consideration are provided in Table~\ref{table:models}.

\begin{table}[b]
    \centering
    \begin{tabular}{*{2}{p{0.2\linewidth}p{\dimexpr0.7\linewidth-2\tabcolsep\relax}}} 
        \hline
        \textbf{Model}  & \textbf{Description} \\
        \hline
        BN & model with batch normalizing transform (ref. Section~\ref{section:batch_normalization}) \\
        \hline
        MN & model with mixture normalizing transform (ref. Section~\ref{section:general}) \\
        \hline
        CN & Context Normalization model which uses CN-Training (ref. Algorithm~\ref{alg:two}) in the training step and CN method (ref. 
 Algorithm~\ref{alg:three}) in the inference step. 
        Two versions of CN are used in the experiments: CN on Patches (CN-Patches) and CN on Channels (CN-Channels). 
        \\
        \hline
        CN+ & Context Normalization model which  uses CN-Training (ref.Algorithm~\ref{alg:two}) in training step and CN+ method (ref. Algorithm~\ref{alg:three}) in inference step \\
        \hline
    \end{tabular}
    \caption{Normalization method used in the experiments}
    \label{table:models}
\end{table}

\indent CN-Patches can be used as layer on Transformer-based architecture~\cite{vaswani2017attention} like Vision Transformer (ViT)~\cite{dosovitskiy2020image}. ViT model takes as input a grid of non-overlapping contiguous image patches. CN-Patches consist of normalizing image patches with a vector $\mu$ and $\sigma$ that are learned from the context identifier embedding (ref. Figure~\ref{fig:cn_layer} and Algorithm~\ref{alg:one}). Images are grouped by context (each context is a component of the mixture of Gaussian distributions), and for each context, a unique vector (context identifier) is associated. The context identifier is first embedded for a better representation, and the resulting vector is used to learn the corresponding parameters $\mu$ and $\sigma$ independently. This process is repeated for each input (image) to the neural network during training. 
CN-Patches is a way of integrating the context information into the image patch normalization process. It allows the image patch representation to be adapted to the context of the image, which improves the overall performance of the model.\newline
\indent CN-Channels is designed to be applied directly to images. The parameters $\mu$ and $\sigma$ are vectors of size the number of channels. They are learned independently according to the context by using context identifiers. CN-Channels incorporates the context identifier into the image normalization process. In this case, the context identifier embedding is used to directly adjust the image representation per channel. This allows the model to adapt the image representation to the context, which improves the overall performance of the model.
%\newline
%
%\newline
%\indent This section summarises the following:
%\begin{itemize}
%    \item Compare CN to BN and MN for small and large learning rates using CNN architecture detailed in Table~\ref{table:cnn_mn}.
%    \item CN using superclass as context.
%    \item Image classification on a blended dataset.
%    \item Image similarity estimation.
%    \item CN in domain adaptation.
%\end{itemize}
\subsection{Datasets}\label{section:datasets}
The experiments in this study utilize several benchmark datasets commonly used in the classification community:
\begin{itemize}
\item CIFAR-10: A dataset with 50,000 training images and 10,000 test images, each of size $32\times32$ pixels, distributed across 10 classes~\cite{cifar10_datasets}.
\item CIFAR-100: Derived from the Tiny Images dataset, it consists of 50,000 training images and 10,000 test images of size $32\times32$, divided into 100 classes grouped into 20 superclasses~\cite{cifar100_datasets}.
\item MNIST digits: Contains 70,000 grayscale images of size $28\times28$ representing the 10 digits, with around 6,000 training images and 1,000 testing images per class~\cite{mnist_datasets}.
\item VERI-Wild: Comprises 416,314 vehicle images with 40,671 identities, captured from different angles and at different times~\cite{lou2019large}.
\item SVHN: A challenging dataset with over 600,000 digit images, focusing on recognizing digits and numbers in natural scene images~\cite{sermanet2012convolutional}.
\end{itemize}
\begin{table}[!h]
    \centering
    \begin{tabular}{*{2}{p{0.2\linewidth}p{\dimexpr0.7\linewidth-2\tabcolsep\relax}}} 
        \hline
        \textbf{Dataset}  & \textbf{Task} \\
        \hline
        CIFAR-10 & This dataset is used for training CIFAR ConvNet (ref. Table~\ref{table:cnn_mn}), ensuring consistent conditions with the study on mixture normalization (ref. Section~\ref{cifar_cnn}). \\
        \hline
        CIFAR-100 & 
        Three experiments (ref. Sections~\ref{cifar_cnn},~\ref{prior_knowledge}, and~\ref{blended_dataset}) use this dataset: comparing with mixture normalization methods, utilizing superclass structure as context, and blending with MNIST digits for domain adaptation. \\
        \hline
        MNIST digits & MNIST digits is employed in two domain adaptation experiments, serving as the blended dataset with CIFAR-100 in the first (ref. Section~\ref{blended_dataset}) and as the source domain in the second (ref. Section~\ref{domain_adaptation}). 
        \\
        \hline
        VERI-Wild & This dataset facilitates the analysis of context parameters obtained through model training using an image similarity measurement strategy (ref. Section~\ref{image_similarity}), thereby constructing the American night. \\
        \hline
        SVHN & SVHN is used in conjunction with MNIST digits as the target dataset to train the AdaMatch model in unsupervised domain adaptation (ref. Section~\ref{domain_adaptation}).\\
        \hline
    \end{tabular}
    \caption{Summary of datasets and associated training experiments}
    \label{table:dataset_task}
\end{table}
Table~\ref{table:dataset_task} illustrates the mapping of each dataset to the corresponding experiments in which it is employed.
\subsection{Context Normalization vs Mixture Normalization on CIFAR}
\label{cifar_cnn}
In this experiment, we use a shallow convolutional neural network architecture as described in Table~\ref{table:cnn_mn}. 
\begin{table}[!ht]
    \centering
    \begin{tabular}{lllll}
        \hline
        layer  & type & size & kernel & (stride, pad) \\
        \hline
        input &  input    & 3 $\times$ 32 $\times$ 32 & \_ & \_ \\
        conv1 & conv+bn+relu & 64 $\times$ 32 $\times$ 32 &  5 $\times$ 5 & (1, 2) \\
        pool1 &  max pool & 64 $\times$ 16 $\times$ 16 & 3 $\times$ 3 & (2, 0) \\
        conv2 & conv+bn+relu & 128 $\times$ 16 $\times$ 16 & 5 $\times$ 5 & (1, 2) \\
        pool2 &  max pool & 128 $\times$ 8 $\times$ 8 & 3 $\times$ 3 & (2, 0) \\
        conv3 & conv+bn+relu & 128 $\times$ 8 $\times$ 8 & 5 $\times$ 5 & (1, 2) \\
        pool3 &  max pool & 128 $\times$ 4 $\times$ 4 & 3 $\times$ 3 & (2, 0) \\
        conv4 & conv+bn+relu & 256 $\times$ 4 $\times$ 4 & 3 $\times$ 3 & (1, 1) \\
        pool4 &  avg pool & 256 $\times$ 1 $\times$ 1 & 4 $\times$ 4 & (1, 0) \\
        linear & linear & 10 or 100 & \_ & \_ \\
        \hline
    \end{tabular}
    \caption{CIFAR ConvNet architecture on mixture normalization paper for a comparison of MN and BN for small and large learning rate regimes}
    \label{table:cnn_mn}
\end{table}

\indent The mixture normalization approach replaces the BN layer in conv3 
with mixture normalization layer, and the context normalization experiment uses CN-Channels as the first CIFAR ConvNet layer, incorporating context information directly into the base image. According to~\cite{kalayeh2019training}, we experiment with two different learning rates, one 5 times larger than the other. The same is done with weight decay. The mini-batch is set to 256, and all models are trained for 100 epochs using RMSprop~\cite{tieleman2017divide} with 0.9 momentum. During training on CIFAR-10 and CIFAR-100, we fit a Gaussian mixture model by Maximum Likelihood Estimation (MLE). We use K-menas++~\cite{arthur2007k} to initialize the centers of the mixture component and Expectation-Maximization (EM)~\cite{Dempster77maximumlikelihood} to estimate the parameters of the mixture model $\theta = \{\lambda_k, \mu_k, \Sigma_k: k = 1, ..., K=3\}$. In mixture normalization, each sample is normalized using the mean and standard deviation of the mixture component to which it belongs. Context normalization treats each Gaussian component as an individual context, enabling sample normalization within each component. It employs separate multi-layer perceptrons (MLPs) to estimate the mean and standard deviation for each context using the one-hot encoded context identifier ($onehot(.)$).\newline
\indent Our primary objective is not to achieve state-of-the-art results, which require computationally expensive architectures and careful parameter tuning. Instead, we aim to demonstrate that by replacing or incorporating our context normalization technique, the convergence rate can be improved, leading to superior test accuracy. This showcases the significant impact of our approach in enhancing model performance.
\begin{table}[htbp]
    \centering
    \begin{tabular}{lll}
        \hline
        \multicolumn{3}{c}{\textbf{CIFAR-10}}\\
        \hline
        model  & (lr, weight decay) & test accuracy (\%) \\
        \hline
        BN-1 & (0.001, 1e-4) & 86.9 \\
        BN-2 & (0.001, 2e-5) & 85.9 \\
        BN-3 & (0.005, 1e-4) & 82.56 \\
        BN-4 & (0.005, 2e-5) & 83.71 \\
        \hline
        MN-1 & (0.001, 1e-4) & 87.08 \\
        MN-2 & (0.001, 2e-5) & 87.9 \\
        MN-3 & (0.005, 1e-4) & 82.07 \\
        MN-4 & (0.005, 2e-5) & 84.71 \\
        \hline
        \textbf{CN-1} & \textbf{(0.001, 1e-4)} & \textbf{87.32} \\
        \textbf{CN-2} & \textbf{(0.001, 2e-5)} & \textbf{88.15} \\
        \textbf{CN-3} & \textbf{(0.005, 1e-4)} & \textbf{83.85} \\
        \textbf{CN-4} & \textbf{(0.005, 2e-5)} & \textbf{87.09} \\
        \hline
    \end{tabular}

        \begin{tabular}{lll}
        \hline
        \multicolumn{3}{c}{\textbf{CIFAR-100}}\\
        \hline
        model  & (lr, weight decay) & test accuracy (\%) \\
        \hline
        BN-1 & (0.001, 1e-4) & 57.48 \\
        BN-2 & (0.001, 2e-5) & 57.69 \\
        BN-3 & (0.005, 1e-4) & 50.82 \\
        BN-4 & (0.005, 2e-5) & 52.25 \\
        \hline
        MN-1 & (0.001, 1e-4) & 61.6 \\
        \textbf{MN-2} & \textbf{(0.001, 2e-5)} & \textbf{61.9} \\
        MN-3 & (0.005, 1e-4) & 53.08 \\
        MN-4 & (0.005, 2e-5) & 52.7 \\
        \hline
        \textbf{CN-1} & \textbf{(0.001, 1e-4)} & \textbf{61.79} \\
        CN-2 & (0.001, 2e-5) & 60.53 \\
        \textbf{CN-3} & \textbf{(0.005, 1e-4)} & \textbf{56.08} \\
        \textbf{CN-4} & \textbf{(0.005, 2e-5)} & \textbf{53.54} \\
        \hline
    \end{tabular}
    \caption{
    Performance Evaluation on CIFAR-10 and CIFAR-100 using the CIFAR ConvNet architecture (ref. Table~\ref{table:cnn_mn}) with the incorporation of Batch Normalization (BN), Mixture Normalization (MN), and Context Normalization (CN). Each algorithm is followed by a corresponding number, representing the  learning rate  and weight decay.
    }
    \label{table:cifar_cnn}
\end{table}
%\indent 
Increasing the learning rate from $0.001$ to $0.005$ increases the convergence gap, showing the ability of context normalization as mixture normalization to take advantage of higher learning rates for training.
\begin{figure}[htbp]
     \centering
     \begin{subfigure}[b]{0.23\textwidth}
         \centering
         \includegraphics[width=\textwidth]{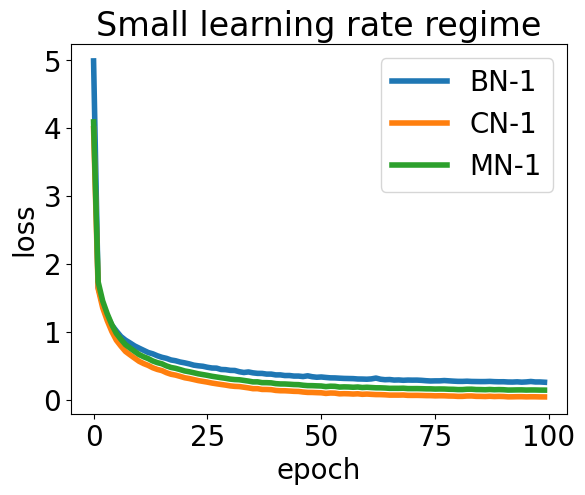}
         \includegraphics[width=\textwidth]{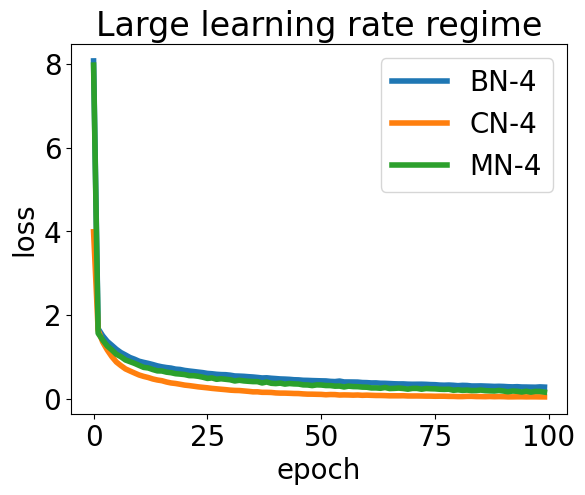}
         \caption{CIFAR-10}
         \label{fig:y equals x}
     \end{subfigure}
     \hfill
     \begin{subfigure}[b]{0.23\textwidth}
         \centering
         \includegraphics[width=\textwidth]{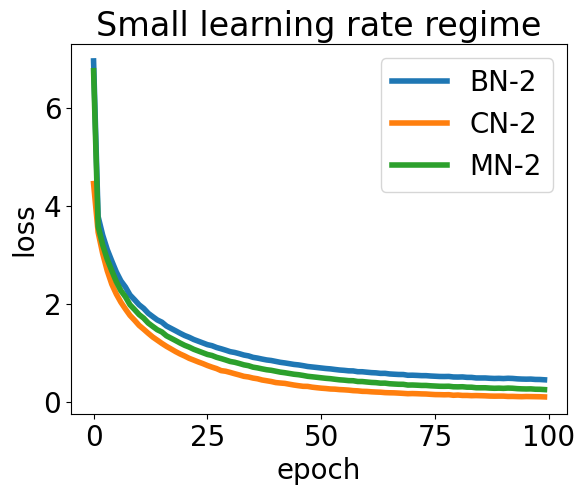}
         \includegraphics[width=\textwidth]{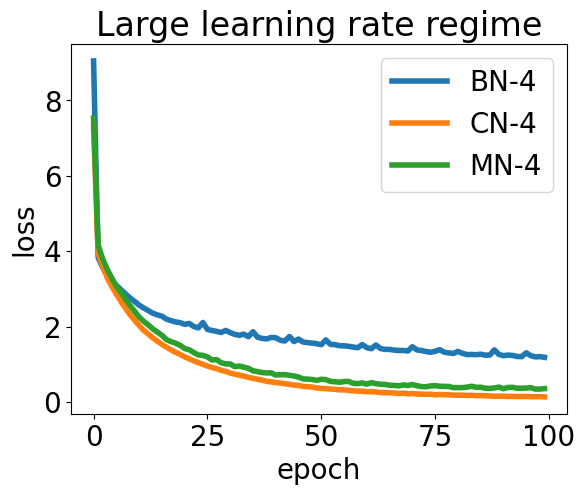}
         \caption{CIFAR-100}
         \label{fig:three sin x}
     \end{subfigure}
        \caption{
         Comparative analysis of validation error curves for the CIFAR ConvNet architecture (ref. Table~\ref{table:cnn_mn}) trained under different learning rate and weight decay configurations.
        %Test error curves when the CIFAR ConvNet architecture (ref. Table~\ref{table:cnn_mn}) is trained under different learning rate and weight decay settings.
        }  
        \label{fig:cifar_cnn_curves}
\end{figure}
From Table~\ref{table:cifar_cnn} and Figure~\ref{fig:cifar_cnn_curves} we can see that regardless of the weight decay and learning rates, CN models not only converge much faster than their corresponding BN and MN counterparts but also achieve better accuracy.\newline
\indent After model training, each context's parameters are known. The use of CN+ in the inference step leads to the results shown in Table~\ref{table:cifar_cnn+}.
\begin{table}[htbp]
    \centering
    \begin{tabular}{lll}
        \hline
        \multicolumn{3}{c}{\textbf{CIFAR-10}}\\
        \hline
        model  & (lr, weight decay) & test accuracy (\%) \\
        \hline
        \textbf{CN+1} & \textbf{(0.001, 1e-4)} & \textbf{87.90} \\
        \textbf{CN+2} & \textbf{(0.001, 2e-5)} & \textbf{88.28} \\
        CN+3 & (0.005, 1e-4) & 82.85 \\
        \textbf{CN+4} & \textbf{(0.005, 2e-5)} & \textbf{87.50} \\
        \hline
    \end{tabular}

        \begin{tabular}{lll}
        \hline
        \multicolumn{3}{c}{\textbf{CIFAR-100}}\\
        \hline
        model  & (lr, weight decay) & test accuracy (\%) \\
        \hline
        CN+1 & (0.01, 1e-4) & 62.01 \\
        CN+2 & (0.001, 2e-5) & 60.53 \\
        CN+3 & (0.005, 1e-4) & 55.80 \\
        CN+4 & (0.005, 2e-5) & 53.15 \\
        \hline
    \end{tabular}
    
    \caption{Performance Evaluation  on CIFAR-10 and CIFAR-100 using CIFAR ConvNet architecture (ref. Table~\ref{table:cnn_mn}). Introducing CN+ which is  an extended version of CN for Inference, encompassing all contexts as Mixture Normalization. Each CN+ variant is denoted by a  number, representing the iterative exploration of learning rate and weight decay adjustments.
    %CN+ is the extended version of CN for inference. It can take into account all contexts as mixture normalization. CN+ is followed by a number. This number indicates the number of the trial by varying the learning rate and weight decay.
    }
    \label{table:cifar_cnn+}
\end{table}
The CN method (ref. Table~\ref{table:cifar_cnn}) is faster than the CN+ method (ref. Table~\ref{table:cifar_cnn+}) and gives approximately the same results. In the following experiments, we use the CN method in the inference step (ref. Algorithm~\ref{alg:three}).\newline
%This experiment is a two-step application of context normalization:
%\begin{itemize}
%    \item Fit a Gaussian mixture model to determine the appropriate components (components are used as contexts in context normalization).
%    \item Assign a context identifier to each context, which will then be used to normalize all the elements that belong to the same context.
%\end{itemize}
\indent In the following experiments, we aim to demonstrate that context normalization can be implemented in a single step in specific scenarios, unlike mixture normalization. This approach would lead to a decrease in time complexity.
%we will show that in some cases, in contrast to the mixture normalization, context normalization can be applied in a single step, thus reducing the time complexity.

\subsection{Context Normalization using superclass as context}\label{prior_knowledge}
As described in Section~\ref{section:datasets}, the CIFAR-100 dataset incorporates a superclass structure in addition to the class partition. Our proposed context normalization leverages this superclass information as "prior knowledge" for classification. Each superclass corresponds to a context, identified by a one-hot encoded vector of size 20 (the number of superclasses). We integrated the context normalization technique into the Vision Transformer (ViT) architecture, using Context Normalization on Patches (CN-Patches) and Context Normalization on Channels (CN-Channels). Training employed early stopping based on validation performance, and images were pre-processed by normalizing them with respect to the dataset's mean and standard deviation. Data augmentation techniques such as horizontal flipping and random cropping were applied to enhance the dataset. The AdamW optimizer with a learning rate of $10^{-3}$ and weight decay of $10^{-4}$ was chosen to prevent overfitting and optimize model parameters~\cite{loshchilov2017decoupled, kingma2014adam}.
\begin{table}[!ht]
    \centering
    \begin{tabular}{lll}
        \hline
        model  & test accuracy & test top-5-accuracy \\
        \hline
        ViT & 52.37\%     & 80.98\% \\
        ViT+BN & 53.35\%    & 79.68\% \\
        \textbf{ViT+CN-Patches} & \textbf{63.80\%}     & \textbf{99.74\%} \\
        \textbf{ViT+CN-Channels} & \textbf{62.48\%}    & \textbf{99.83\%} \\
        \hline
    \end{tabular}
    \caption{Comparison of the two Context Normalization methods on CIFAR-100: Context Normalization on Patches (CN-Patches) and Context Normalization on Channels (CN-Channels), with normalization to the mean and standard deviation of the dataset (ViT) and input normalization using batch normalization (BN).}
    \label{table:cifar100}
\end{table}

\begin{figure*}[!ht]
     \centering
     \begin{subfigure}[b]{0.3\textwidth}
         \centering
         \includegraphics[width=\textwidth]{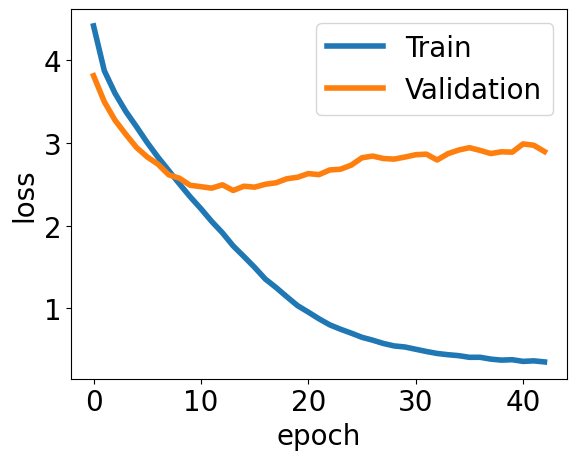}
         \caption{BN}
         \label{fig:bn}
     \end{subfigure}
     \hfill
     \begin{subfigure}[b]{0.3\textwidth}
         \centering
         \includegraphics[width=\textwidth]{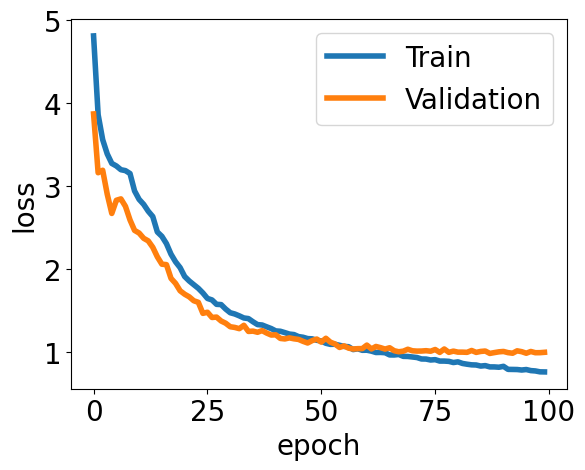}
         \caption{CN-Patches}
         \label{fig:CN-Patches}
     \end{subfigure}
     \hfill
     \begin{subfigure}[b]{0.3\textwidth}
         \centering
         \includegraphics[width=\textwidth]{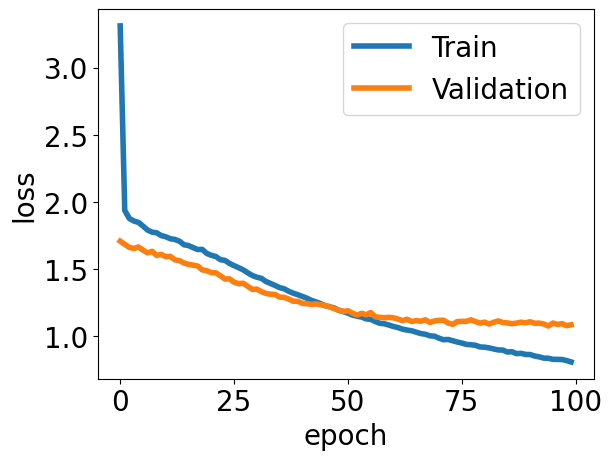}
         \caption{CN-Channels}
         \label{fig:CN-Channels}
     \end{subfigure}
        \caption{Validation loss curves on CIFAR-100 when the ViT architecture is trained with different normalization methods.}
        \label{fig:cifar100_loss}
\end{figure*}

Table~\ref{table:cifar100} demonstrates the significant performance improvement of context normalization over batch normalization (BN) when using the ViT architecture trained from scratch on CIFAR-100. Both CN-Patches and CN-Channels approaches outperform BN by approximately 10\% and 18\% in terms of accuracy and top-5 accuracy. The train and test loss comparison in Figure~\ref{fig:cifar100_loss} further supports this observation, showing that CN-Patches and CN-Channels accelerate the learning process and enhance classification performance. These results indicate that the proposed approaches stabilize the distribution and mitigate internal covariate shift, resulting in a superior data representation aligned with the target task.
\subsection{Image classification on blended dataset}\label{blended_dataset}
As deep neural networks require a certain amount of labeled data for effective training, it is well known that the lack of a large enough corpus of accurately labeled high-quality data can produce disappointing results. Data augmentation~\cite{perez2017effectiveness} is one way to overcome this problem. However, current approaches generate the data with a distribution that may be quite different from the original one, and this data bias will often lead to suboptimal performance~\cite{xu2020wemix}. Inspired by the transfer learning~\cite{torrey2010transfer}, we study in this subsection a new approach using the proposed context normalization. The goal of this approach is to improve the performance of the model on a target dataset by training the model on the combined dataset and then applying the trained model to the target dataset. In this framework, the proposed normalization technique allows to obtain a domain adaptation.   

Specifically, we train ViT models with the same settings as in Section~\ref{prior_knowledge} with context normalization approaches (CN-Channels and CN-Patches) on the combined dataset CIFAR-100 and MNIST digits. We target two contexts $r \in \{1, 2\}$, corresponding to the datasets and the context identifier is encoded by $onehot(r=1)=(1,0)$ for images in CIFAR-100 and $onehot(r=2)=(0,1)$ for images in MNIST digits. The parameters $\mu$ and $\sigma$ of each Gaussian distribution are then learned using these vectors after embedding. The trained models (on the blended dataset) are then applied to the CIFAR-100 test dataset.

It is important to mention that the baseline models (ViT with standard preprocessing and ViT with batch normalization) collapsed in this blended dataset as the two datasets have different structures, and simple normalization does not allow a suitable representation of the data. Context normalization, on the other hand, gives an adaptive representation per dataset (according to the contexts), which makes training possible. As shown in Table~\ref{table:cifar100+mnist}, models with context normalization technique achieve good results on the blended dataset. It is also interesting to notice that this performance in terms of accuracy is biased by the MNIST digits dataset, which is less difficult to learn than CIFAR-100. More precisely, the model with CN-Patches achieves 55.04\% accuracy and 81.83\% top-5 accuracy, which exceeds the results of all baseline models (ref. Figure~\ref{table:cifar100}) trained on CIFAR-100. The model with CN-Channels gives 50.99\% accuracy and 78.55\% top-5 accuracy.
\begin{table}[!ht]
    \centering
    \begin{tabular}{lll}
        \hline
        model  & accuracy & top-5-accuracy \\
        \hline
        ViT &  \_    & \_ \\
        ViT+BN & \_ & \_ \\
        \textbf{ViT+CN-Patches} & \textbf{77.09\%}     & \textbf{90.92\%} \\
        \textbf{ViT+CN-Channels} & \textbf{74.92\%}    & \textbf{89.28\%} \\
        \hline
    \end{tabular}
    \caption{Blended dataset CIFAR-100 and MNIST digits: results of models based on the two normalization methods (CN-Patches and CN-Channels). Baseline models collapsed in this combined dataset.}
    \label{table:cifar100+mnist}
\end{table}

The Latent space normalization (CN-Patches) has a better performance than the CN-Channels in this experiment. The drop in performance for the CN-Channels method can be explained by the fact that the normalization is applied directly to images located in a space with several different data characteristics.
\subsection{Self-supervised learning for image similarity estimation}\label{image_similarity}
This experiment is motivated by the interpretability of context normalization parameters of each context obtained after training. To illustrate this, we construct the American night with context normalization on the VERI-Wild dataset. American night~\cite{haro2006visual} is a set of cinematic techniques used to simulate a night scene while filming in daylight. To accomplish this, we make a partition of two contexts on the dataset: $r \in \{Day, Night\}$. A context vector $onehot(r=Day)=(1,0)$ is assigned to day images and $onehot(r=Night)=(0,1)$ to night images. We label pairs of images containing the same object with 1, and otherwise with 0.

\begin{figure}[htbp]
     \centering
     \begin{subfigure}[b]{0.23\textwidth}
         \centering
         \includegraphics[width=\textwidth]{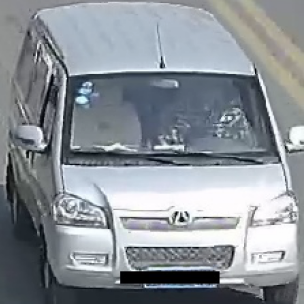}
         \caption{Daytime image}
         \label{fig:y equals x}
     \end{subfigure}
     \hfill
     \begin{subfigure}[b]{0.23\textwidth}
         \centering
         \includegraphics[width=\textwidth]{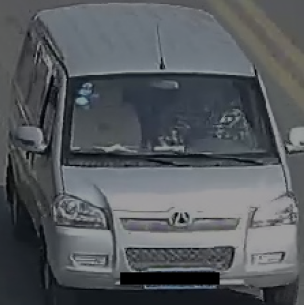}
         \caption{Nighttime image simulation}
         \label{fig:three sin x}
     \end{subfigure}
        \caption{
Simulation of a night image on a day image: night image is obtained by normalizing day image with Day parameters, then scale and shift with Night parameters.
        }  
        \label{fig:day}
\end{figure}

\begin{figure}[htbp]
     \centering
     \begin{subfigure}[b]{0.23\textwidth}
         \centering
         \includegraphics[width=\textwidth]{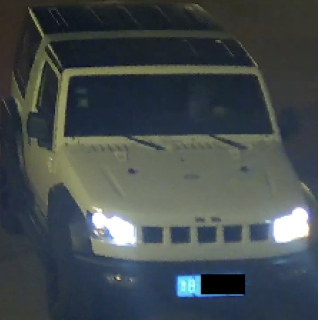}
         \caption{Nighttime image}
         \label{fig:y equals x}
     \end{subfigure}
     \hfill
     \begin{subfigure}[b]{0.23\textwidth}
         \centering
         \includegraphics[width=\textwidth]{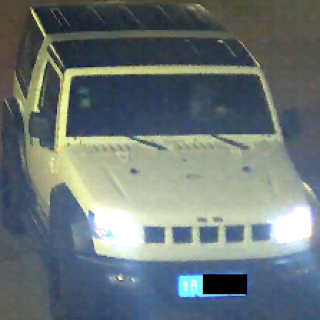}
         \caption{Daytime image simulation}
         \label{fig:three sin x}
     \end{subfigure}
        \caption{
Simulation of day image on night image: day image is obtained by normalizing night image with Night parameters, then scale and shift with Day parameters
        }  
        \label{fig:night}
\end{figure}

We use context normalization on the backbone of a siamese network (ViT architecture) with contrastive loss~\cite{hadsell2006dimensionality} to estimate the similarity between images.
The aim of this experiment is to reveal the behavior of the parameters learned by the model and to understand how context information has an influence on the normalization process. The use of a Siamese network makes it possible to measure the similarity between the images and to evaluate the efficiency of the normalization in preserving the relevant information.\newline
After training, we normalize day images with the appropriate parameters ($\mu_{Day}$ and $\sigma_{Day}$), then denormalize (scale and shift) with the night parameters ($\mu_{Night}$ and $\sigma_{Night}$), to construct night images, as shown in Figure~\ref{fig:day}. The images become darker, giving the night effect known as the American night (day for night) used for cinematic production in Hollywood, where a filter is applied to emphasize the light in the blue channel. A reverse process can be applied to night images to obtain day images, as shown in Figure~\ref{fig:night}, with brighter images reflecting the daylight effect.  

\subsection{Context normalization in domain adaptation}\label{domain_adaptation}
In this experiment, we show that due to its strength in local representation, context normalization can yield substantial gains in domain adaptation. Domain adaptation~\cite{farahani2021brief} is a technique for using knowledge learned by a model from a related domain with sufficient labeled data to improve the performance of the model in a target domain with insufficient labeled data. In this case, two contexts can be considered: $r \in $ \{source domain, target domain\}. As an illustration, we use context normalization with AdaMatch~\cite{berthelot2021adamatch}, a method that combines the tasks of unsupervised domain adaptation (UDA), semi-supervised learning (SSL) and semi-supervised domain adaptation (SSDA). In UDA, we have access to a source labeled dataset and a target unlabeled dataset. Then the task is to learn a model that can generalize well to the target dataset. The source and the target datasets vary in terms of distribution. We use the MNIST dataset (ref. Section~\ref{section:datasets}) as the source dataset, while the target dataset is SVHN (ref. Section~\ref{section:datasets}). Both datasets have various varying factors in terms of texture, viewpoint, appearance, etc. Their domains, or distributions, are different from one another. As in~\cite{berthelot2021adamatch}, we use wide residual networks~\cite{zagoruyko2016wide} for the dataset pairs. The model is trained using the Adam~\cite{kingma2014adam} optimizer and a cosine decay schedule to reduce the initial learning rate, which is initialised at 0.03. Context normalization is used as the first layer of AdaMatch to incorporate the context identifier (source domain and target domain) into the image normalization process.
\begin{table}[!ht]
    \centering
    \begin{tabular}{lll}
    \hline
    model  & source data (MNIST) & target data (SVHN) \\
    \hline
     AdaMatch &  79.39\%    & 20.46\% \\
     \textbf{AdaMatch+CN-Channels} & \textbf{99.21\%} & \textbf{43.80\%} \\
    \hline
    \end{tabular}
    \caption{Test accuracy of AdaMatch and AdaMatch with context normalization (AdaMatch+CN-Channels) using source domain (MNIST) as context identifier.} 
    \label{table:source_domain}
\end{table}

\begin{table}[!ht]
        \centering
        \begin{tabular}{lll}
        \hline
        model  & source data (MNIST) & target data (SVHN) \\
        \hline
         AdaMatch &  79.39\%    & 20.46\% \\
         \textbf{AdaMatch+CN-Channels} & \textbf{94.45\%}  & \textbf{23.22\%} \\
        \hline
        \end{tabular}
        \caption{Test accuracy of AdaMatch and AdaMatch with context normalization (AdaMatch+CN-Channels) using target domain (SVHN) as context identifier.} 
        \label{table:target_domain}
\end{table}
In general, the clear improvement that context normalization brings in terms of validation can be seen in Tables~\ref{table:source_domain} and~\ref{table:target_domain}. Normalizing with CN-Channels takes the target task into account. The source domain (MNIST) is labeled as opposed to the target domain (SVHN), which could explain why using MNIST as the context identifier in AdaMatch+CN-Channels (ref. Tables~\ref{table:source_domain} and~\ref{table:target_domain}) gives better results.
\section{Conclusion}
We have proposed a novel approach called "context normalization" (CN) that enhances deep neural network training in terms of training stability, fast convergence, higher learning rate, and viable activation functions. Similar to the conventional mixture normalization (MN) method, our approach is driven by the hypothesis that any continuous function can be approximated in some sense by a weighted sum of Gaussian distributions with finite mean vectors and covariance matrices. In other words, our methodology assumes that the data distribution is a mixture of Gaussian models. However, unlike the mixture normalization technique that invokes the expectation maximization (EM) algorithms to estimate the Gaussian components parameters, our proposed methodology relies on the notion of concept that represents a cluster of related data. In fact, a supervised deep neural network is built and trained in order to learn the Gaussian components parameters. Once these optimal values are determined after convergence, they are utilized during the CN procedure performed on a deep neural network activation layer. CN alleviates the slow estimation of Gaussian component parameters inherent to EM in the scenario of large datasets. Furthermore, unlike MN, CN provides non linear decision boundaries between context which reflects more reality. Our experimental results demonstrate the superiority of context normalization over batch normalization and mixture normalization, showcasing enhanced convergence and generalization performance. The proposed method, when applied specifically to images, introduces CN-Channels and CN-Patches for training, and CN and CN+ for inference. With its flexibility to adapt various representations and tasks, context normalization proves to be a valuable tool in some application such as image classification.\newline
\indent Our short-term perspective consists in merging seamlessly a gradient free optimization algorithm with a gradient-based error optimizer in order to reach global convergence. We believe that this objective will boost training of deep neural networks further. This precision margin gained allows gaining insight into the neuron activation level sensitivity.
\bibliographystyle{IEEEtran}
\bibliography{sample-base}
\end{document}